\RequirePackage{xcolor}
\documentclass[journal,twoside,web]{ieeecolor}

\usepackage{algorithm,algorithmicx,algpseudocode}
\usepackage{tmi}
\usepackage{cite}
\usepackage{array}
\usepackage{amsmath,amssymb,amsfonts,mathtools}
\usepackage{graphicx}
\usepackage{textcomp}
\usepackage{url}

\def\BibTeX{{\rm B\kern-.05em{\sc i\kern-.025em b}\kern-.08em
    T\kern-.1667em\lower.7ex\hbox{E}\kern-.125emX}}
\begin{document}
\title{DiffBoost: Enhancing Medical Image Segmentation via Text-Guided Diffusion Model}
\author{Zheyuan Zhang, Lanhong Yao, Bin Wang,  Debesh Jha, Gorkem Durak, Elif Keles, Alpay Medetalibeyoglu,  Ulas Bagci
\thanks{This work was supported by the NIH NCI R01-CA246704 and NIH/NIDDK \#U01 DK127384-02S1.}
\thanks{All authors belong to Machine \& Hybrid Intelligence Lab in Northwestern University, Chicago IL 60611, USA}
\thanks{Corresponding author: Ulas Bagci (ulas.bagci@northwestern.edu)}}

\maketitle

\begin{abstract}
Large-scale, big-variant, high-quality data are crucial for developing robust and successful deep-learning models for medical applications since they potentially enable better generalization performance and avoid overfitting. However, the scarcity of high-quality labeled data always presents significant challenges. This paper proposes a novel approach to address this challenge by developing controllable diffusion models for medical image synthesis, called \textcolor{black}{DiffBoost}. We leverage recent diffusion probabilistic models to generate realistic and diverse synthetic medical image data that preserve the essential characteristics of the original medical images by incorporating edge information of objects to guide the synthesis process. In our approach, we ensure that the synthesized samples adhere to medically relevant constraints and preserve the underlying structure of imaging data. Due to the random sampling process by the diffusion model, we can generate an arbitrary number of synthetic images with diverse appearances. To validate the effectiveness of our proposed method, we conduct an extensive set of medical image segmentation experiments on multiple datasets, including Ultrasound breast (+13.87\%), CT spleen (+0.38\%), and MRI prostate (+7.78\%), achieving significant improvements over the baseline segmentation methods. The promising results demonstrate the effectiveness of our \textcolor{black}{DiffBoost} for medical image segmentation tasks and show the feasibility of introducing a first-ever text-guided diffusion model for general medical image segmentation tasks. With carefully designed ablation experiments, we investigate the influence of various data augmentations, hyper-parameter settings, patch size for generating random merging mask settings, and combined influence with different network architectures. Source code with checkpoints are available at  \url{https://github.com/NUBagciLab/DiffBoost}.


\end{abstract}


\section{Introduction}
\label{sec1}
\textcolor{black}{The recent surge of artificial intelligence (AI) / deep learning in medical image analysis has revolutionized the field. However, achieving optimal performance with these techniques often hinges on access to large-scale, high-quality annotated datasets. Unfortunately, the medical domain faces a significant challenge – \textit{data scarcity}. This scarcity stems from several factors: \textbf{(i)} acquiring and annotating high-quality medical images is a time-consuming, labor-intensive, and expensive endeavor.  \textbf{(ii) }Collaboration in data sharing and collection can be limited by privacy concerns and ethical considerations. \textbf{(iii)} The challenge is amplified in the context of rare diseases, where data availability is even more limited.}

\textcolor{black}{The data scarcity is surely a bottleneck, impeding the full potential of deep learning in medical image analysis. Hence, the medical imaging community continuously explores innovative solutions to overcome this hurdle and unlock further advancements. Some potential solutions for this important problem are the following: data augmentation, transfer learning and domain adaptation, federated learning, and lightweight deep learning architectures. In this study, our proposed strategy for getting larger and more diverse data is based on a new data augmentation strategy with a generative AI model, specifically a stable-diffusion model, and we apply a downstream task of segmentation to explore its efficacy on one of the most important medical image analysis tasks - segmentation.}

Data augmentation is a widely adopted approach to alleviate the issue of limited annotated data by expanding the training dataset by generating new samples~\cite{chlap2021reviewmedaug}. This fosters improved learning and generalization capabilities in deep learning models by enriching the dataset with diverse and representative examples and mitigating the risk of overfitting. Data augmentation techniques can be broadly categorized into two groups: transformation-based methods and generative methods. Both categories aim to expand the available training data by generating new samples that maintain the essential visual clues of the original data.

Transformation-based data augmentation methods involve the application of basic transformations to the original data samples~\cite{van2001artaug}. These techniques are relatively simple and computationally efficient. Some common traditional methods include spatial level: Rotation, Scaling, Translation, Flipping, and Intensity level: Contrast Adjustment and gamma Correction. Traditional data augmentation can provide explainable and reliable augmented data with low computation costs, which has been widely accepted as an essential element in medical AI applications~\cite{zhang2020deepaug,isensee2021nnunet}. While traditional methods are easy to implement and computationally less demanding, they may not be sufficient to capture the complex variations and high-dimensional relationships present in medical images, particularly in the context of diverse patient populations~\cite{garcea2022medaugreview}.

Generative data augmentation methods, on the other hand, employ advanced models, such as adversarial image attacking and other generative AI models to synthesize more complex and realistic synthetic samples~\cite{shin2018advermedical}. These methods offer the potential to capture the intricate variations and relationships within medical imaging data. \textbf{(i) }Adversarial attack-based data augmentation involves generating perturbed versions of the original images that can deceive the model into making incorrect predictions. These perturbations are carefully designed to maintain the visual appearance of the original images while causing the model to misclassify them~\cite{goodfellow2014advernoise,zhang2023adverin,chen2020realistic}. \textbf{(ii)} Generative AI models: Until recently, the most widely used generative model for the medical imaging was the Generative Adversarial Networks (GANs). GANs consist of two neural networks, a generator, and a discriminator trained in tandem. The generator creates synthetic images, while the discriminator assesses the realism of these images by comparing them to the original data. Through this adversarial process, the generator learns to produce increasingly realistic samples, which can be utilized for data augmentation. It is also worth noting that most adversarial attack models conduct data augmentation during segmentation training, while the generative models are trained before training the segmentation.

More recently, \textit{denoising diffusion probabilistic models (DDPM)} represent a novel topic in generative AI, showing impressive performance in high-quality image synthesis~\cite{ho2020ddpm,rombach2022stablediffusion}, surpassing GAN models. By simulating a stochastic reverse diffusion process, the diffusion models gradually transform an initial noise sample into a realistic data sample through a series of denoising steps. \textcolor{black}{The forward process in diffusion models gradually adds noise to the data, step by step, until the data (image) becomes pure noise.} The main goal of the diffusion model is to learn the reverse process to denoise the corrupted samples and recover the original data~\cite{ho2020ddpm,chen2020realistic}. By leveraging a denoising score-matching objective, the model learns a denoising function that estimates the gradient of the data distribution's log density. The sampling process starts with an initial noise sample and iteratively refines it by applying the learned denoising function at each time step $t$, following the inverse noise schedule. At each step, the model estimates the gradient of the log-density and updates the current sample accordingly. After a predefined number of reverse diffusion steps, an initial noise sample is iteratively refined according to the learned diffusion process, resulting in high-quality synthetic samples that resemble the original data~\cite{ho2020ddpm}.

Compared with the GAN methods, diffusion models have several advantages~\cite{croitoru2023diffusionsurvey}. First, GANs are known for their training instability, which arises from the adversarial min-max optimization process between the generator and discriminator networks. This can lead to issues such as mode collapse and vanishing gradients. In contrast, diffusion models employ a denoising score-matching objective,  a more stable and well-behaved optimization problem. This leads to more stable training and better convergence. Second, diffusion models have demonstrated the ability to generate high-quality samples with sharp and detailed features. This is critical for medical image analysis tasks, where the generated samples are in need to be both visually realistic and medically relevant. GANs, on the other hand, can sometimes produce samples with noticeable artifacts or unrealistic features. Third, in diffusion models, the sampling process is performed through a series of reverse diffusion steps that gradually refine an initial noise sample. This process can be controlled by adjusting the number of steps, \textit{noise schedule}, and other parameters, allowing for fine-grained control over the generated samples. This is essential for generating samples with varying degrees of complexity and diversity. In contrast, GANs typically provide less explicit control over the sampling process. Finally, due to the denoising objective, diffusion models can be more robust to overfitting than GANs and avoid mode collapse. This is particularly important when working with limited data, as is often the case in medical image analysis tasks~\cite{dhariwal2021diffbeatgan}.

\textcolor{black}{In this study, we propose a text-guided diffusion model-based (DDPM) data augmentation approach, called \textcolor{black}{DiffBoost}, to enhance the performance of downstream medical image segmentation tasks by generating reliable and medically relevant synthetic images to be used in training of segmentation algorithm.} The proposed approach consists of the following steps: 
 \textbf{i)} \textbf{Pretraining on large medical datasets:} We begin by pretraining a diffusion model on \textit{RadImageNet} \cite{mei2022radimagenet}, a comprehensive medical image dataset. \textbf{ii)} \textbf{Fine-tuning on downstream task:} Following the pretraining, the diffusion model is fine-tuned on a smaller dataset specific to the target downstream task (segmentation). This adaptation process allows our model to account for unique characteristics and variations present in the task-specific data, resulting in more relevant synthetic samples for data augmentation. \textbf{iii)} \textbf{Integration with Downstream Task Training:} \textcolor{black}{To enrich our training data, we leverage a fine-tuned diffusion model to generate new samples. These synthetic samples incorporate text and edge information for guidance. During model training, a carefully designed combination loss ensures both real and synthetic samples contribute equally, fostering better generalization for the target task.}

We validate the effectiveness of the proposed method DiffBoost on various datasets, including breast ultrasound, spleen CT, and prostate MRI, with extensive experiments. \textcolor{black}{We employ ablation experiments for a systematic exploration of the impact of data augmentation ratios, hyperparameter tuning, patch size selection within the random merging mask generation process, and the interplay between different network architectures. Through this analysis, we gain valuable insights into the individual and combined contributions of these components to model performance on segmentation tasks.} 

\section{Related Work}

\textbf{Stable Diffusion Model}: Rombach et al.~\cite{rombach2022stablediffusion} applies the diffusion model in the latent space of pre-trained autoencoders, enabling training on limited computational resources while retaining their quality and flexibility. Cross-attention layers turn diffusion models into powerful, flexible generators with various conditioning inputs, such as text or bounding boxes. The CLIP~\cite{radford2021clip} builds a strong connection between the image and text representation through large-scale text-image pair training. The combination of CLIP and Stable diffusion enables us to directly generate high-quality and meaningful images from the text, as shown in~\cite{rombach2022stablediffusion}. Based on the pre-trained large-scale text-to-image stable diffusion model, many interesting applications have been proposed in computer vision, like Textual Inversion~\cite{gal2022textinverse}, Subject Driven Generation~\cite{ruiz2022dreambooth}, Pix2Pix Text Instruct Translation~\cite{brooks2022instructpix2pix}. \textcolor{black}{Building upon existing text-to-image diffusion models,~\cite{zhang2023conditiondiff} incorporated additional information like edge maps, segmentation information, and key points. This allows for greater control over the generated image, ensuring it reflects the provided structural details through the conditioning process.}

\textbf{Diffusion Models in Medical Applications}: Diffusion models have emerged as a powerful generative approach for medical image synthesis and augmentation. This innovative technique offers a promising avenue for enhancing the performance of diverse medical image analysis tasks~\cite{kazerouni2022diffusioninmed}. These models have been applied to various data modalities already, from CT to MRI and from 2D to 3D images \cite{jiang2023meddiffmaskgnmri,peng2022meddiffmaskgn3dmri,iskandar2023meddiffmaskgnus,machavcek2023meddiffmaskgnply,muller2022meddiffmaskgnbeatsgan,bedel2023dreamr}. Diffusion models are also used for image-to-image translation applications such as generating CT scans from MRIs \cite{li2023medtranszero,lyu2022medtransctmri,ozbey2022medtransadver}. The diffusion models are used to extract meaningful deep representation features through the reconstruction process too \cite{xu2023openseg,baranchuk2022labelefficient}. The conclusions of such studies show that diffusion algorithms can be useful for medical image segmentation applications, as evidenced in \cite{wu2023medsegdiff,bieder2023medsegeff,rahman2023medsegambiguous}. \textcolor{black}{For a popular topic in brain imaging, authors in 
\cite{dorjsembe2023conditional} introduce Med-DDPM for 3D Brain MRI synthesis, further helping with improving segmentation accuracy from 65.31\% to 66.75\% in terms of Dice score.}

\textcolor{black}{In a very recent work by~\cite{shao2023diffuseexpand}, authors introduce high-quality sampling to enhance the effectiveness of data augmentation. Beyond conventional use, diffusion models are applied to medical image reconstruction~\cite{cui2022meddiffrecon}, denoising~\cite{xiang2023ddm}, and anomaly detection~\cite{wolleb2022diffusion,bercea2023maskdiff}. Last, but not least, authors in \cite{dorjsembe2024polyp} introduce Diffusion-Based semantic polyp synthesis, enhancing polyp segmentation models to be comparable with real endoscopic images. While existing work explores medical image augmentation, leveraging text-based guidance with a pixel-level aligned diffusion model remains under investigated. Our work tackles this gap by exploring this approach from various perspectives.}

\section{Methods}
\begin{figure*}[h]
\centering
\includegraphics[width=0.85\textwidth]{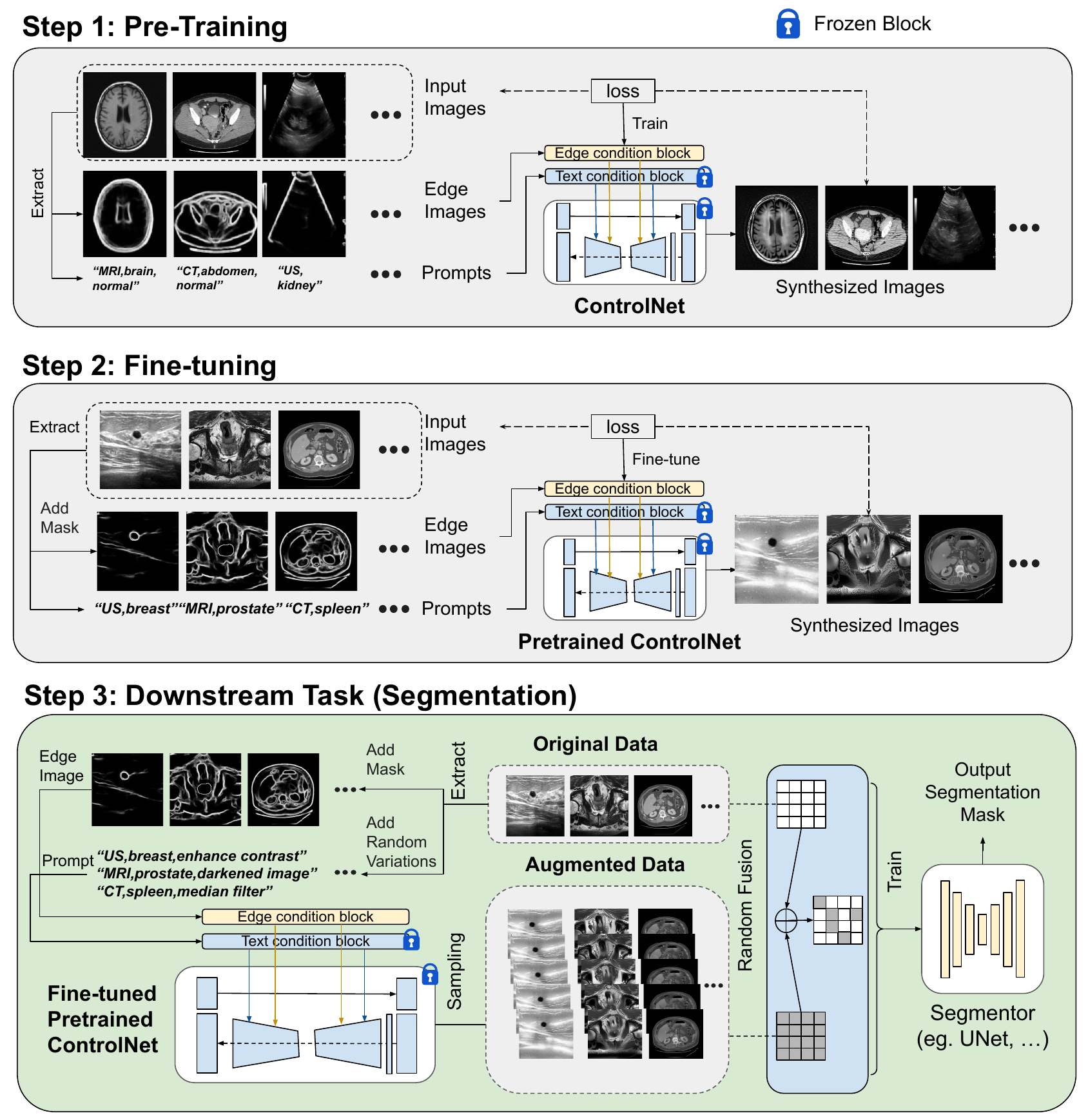}
\caption{Our proposed approach involves three stages: (a) training a diffusion model on a comprehensive radiology imaging dataset (\textit{RadImageNet}), (b) fine-tuning the pre-trained model on a task-specific dataset, allowing for adaptation to the unique characteristics of each target task, and (c) utilizing the fine-tuned model for downstream task training, integrating the synthetic samples (generated during data augmentation) to enhance generalization and performance in the target task (segmentation).}
\label{fig:networkstructure}
\end{figure*}

In this work, we present a text guided diffusion based medical image data augmentation approach, \textit{DiffBoost}, aimed at generating reliable and medically relevant synthetic imaging data to improve the performance of medical image segmentation. Three major steps of our algorithm are illustrated in Figure~\ref{fig:networkstructure}. After introducing the background of DDPM, each of these steps is described below in detail.


\subsection{Denoising Diffusion Probabilistic Models}
Denoising Diffusion Probabilistic Models (DDPMs) define the forward noise process $q$ with $x_0 \sim q(x_0)$ representing the target data distribution, which produces latent $x_1$ through $x_T$ by adding Gaussian noise at time $t$ as follows:
\begin{alignat}{2}
    q(x_1, ..., x_T | x_0) &\coloneqq \prod_{t=1}^{T} q(x_t | x_{t-1}), \label{eq:joint} \\
    q(x_t | x_{t-1}) &\coloneqq \mathcal{N}(x_t; \sqrt{1-\beta_t} x_{t-1}, \beta_t \mathbf{I}), \label{eq:singlestep}
\end{alignat}
where $\beta_t \in (0,1)$ represents the variance schedule across diffusion steps and $\mathbf{I}$ is the identity matrix. With $\alpha_t \coloneqq 1 - \beta_t$ and $\bar{\alpha}_t \coloneqq \prod_{s=0}^{t} \alpha_s$, we can readily sample an arbitrary step of noised latent samples from the $x_0$ according to Equation \ref{eq:singlestep}:
\begin{equation}
    x_t =  \sqrt{\bar{\alpha}_t} x_0 + \sqrt{1-\bar{\alpha}_t} \epsilon. \label{eq:jumpnoise}
\end{equation}
When $T$ is sufficiently large according to the schedule of $\beta_t$, $\bar{\alpha}$ will be nearly 0, and $x_t$ will approximately follow the isotropic Gaussian Distribution. Thus, our goal is to approximate the reverse process $q(x_{t-1}|x_t)$ such that we can use such process to generate a sample $q(x_0)$ from $x_T \sim \mathcal{N}(0, \mathbf{I})$. However, since $q(x_{t-1}|x_t)$ is not directly tractable, we introduce a neural network $\mu_{\theta}(x_t, t)$ to approximate it as follows:
\begin{alignat}{2}
    p(x_0, ..., x_{T-1} | x_T) &\coloneqq \prod_{t=1}^{T} p(x_{t-1} | x_t) \label{eq:reversejoint}, \\
    p_{\theta}(x_{t-1}|x_t) &\coloneqq \mathcal{N}(x_{t-1}; \mu_{\theta}(x_t, t), \Sigma_{\theta}(x_t, t)) .\label{eq:nn}
\end{alignat}
While this approach is promising, Ho et al. \cite{ho2020ddpm} shows that direct parameterization of $\mu_{\theta}(x_t, t)$ may lead to a worse performance. Instead,  predicting a noise $\epsilon$ using a noise prediction network $\epsilon_\theta(x_t, t)$ and generating $\mu_{\theta}(x_t, t)$ may be a better solution:
\begin{equation}
\mu_{\theta}(x_t, t) = \frac{1}{\sqrt{\alpha_t}} \left( x_t - \frac{\beta_t}{\sqrt{1-\bar{\alpha}_t}} \epsilon_{\theta}(x_t, t) \right).
\end{equation}

To train this noise prediction network, we optimize the variational lower bound (VLB) on the negative log-likelihood as follows:
\begin{alignat}{2}
    L_{\text{vlb}} &\coloneqq L_0 + L_1 + ... + L_{T-1} + L_T, \label{eq:loss}, \\
    L_{0} &\coloneqq -\log p_{\theta}(x_0 | x_1), \label{eq:loss0} \\
    L_{t-1} &\coloneqq D_{KL} \left( q(x_{t-1}|x_t,x_0), p_{\theta}(x_{t-1}|x_t) \right), \label{eq:losst} \\
    L_{T} &\coloneqq D_{KL} \left( q(x_T | x_0),  p(x_T) \right),\label{eq:lossT} 
\end{alignat}
where $D_{KL}$ represents the \textit{KL} divergence between two Gaussian distributions. By reparameterizing Equation \ref{eq:loss}, \cite{ho2020ddpm} proposes a simplified objective as follows:
\begin{equation}
    L_{\text{simple}} = E_{t,x_0,\epsilon}\left[ || \epsilon - \epsilon_{\theta}(x_t, t) ||^2 \right], \label{eq:loss_simple}
\end{equation}
where J. Song et al. \cite{song2020scorediff} demonstrates that this loss objective connects with score-based generative networks. Following this training setup, R. Rombach et al. \cite{rombach2022stablediffusion} conducts the diffusion training in latent feature space $z$ where $z=E(x),  \tilde{x}=D(z)$ and $E, D$ represents the encoder and decoder in autoencoder setup, respectively. This powerful design enables us to sample high-quality and high-resolution images within a reasonable time without extensive computation in generating large-size images directly. The capabilities of diffusion models are enhanced as powerful and flexible generators $f$ by incorporating cross-attention layers from various conditioning inputs, such as text or bounding boxes, into the model architecture for controling the generation steps. The $\epsilon_{\theta}(x_t, c, t)$  is optimized according to the objective loss in Equation \ref{eq:loss_simple}, where $c$ represents various conditioning inputs such as the text prompts with CLIP encoding \cite{radford2021clip}. 

In this work, we design the noise prediction $\epsilon_{\theta} (x_t, c_{t}, c_{e}, t)$  network with two types of conditioning input: text $c_{t}$ and edge information $c_{e}$. Our detailed network architecture is shown in Figure~\ref{fig:networkstructure}. Leveraging the benefits of large-scale pre-trained text-to-image stable diffusion models while mitigating unnecessary computational resource burden, we implement a branch design comprising the original branch for stable diffusion with text conditioning input and an auxiliary branch incorporating additional text conditioning input. To ensure the generation of grayscale medical images as opposed to RGB images, we incorporate a cross-channel average block in the final output stage of the model (Figure~\ref{fig:networkstructure}a). \textcolor{black}{With this novel approach, we enable grayscale image generation without requiring a complete network re-architecture. This design optimization streamlines the training process, leading to efficient resource allocation and ultimately culminating in improved overall model performance.}

\subsection{Pre-training on \textit{RadImageNet}}
\textcolor{black}{Currently, most text-to-image diffusion models rely on vast datasets of natural images for training. This lack of focus on large medical datasets presents a clear gap that needs to be addressed. The significant disparity between diffusion models trained on natural images and those designed for medical images presents a critical challenge. Our research addresses this gap by proposing the adaptation of well-trained natural image diffusion models for application in the medical field.} \textcolor{black}{In other words, we first train (fine-tune) the diffusion model on a large-scale medical image dataset, \textit{RadImageNet}, using pre-trained checkpoints from stable diffusion algorithm rather than training from scratch~\cite{zhang2023conditiondiff}.} This pre-trained model serves as a foundation for subsequent adaptation to segmentation tasks, effectively bridging the gap between the two domains. \textit{RadImageNet} is a comprehensive, large-scale medical imaging dataset comprising a diverse array of images from various modalities, such as magnetic resonance imaging (MRI), computed tomography (CT), and ultrasound (US) \cite{mei2022radimagenet}. \textit{RadImageNet} encompasses 11 anatomical regions, including CT scans of the chest, abdomen, and pelvis, MRI scans of the ankle, foot, knee, hip, shoulder, brain, spine, abdomen, and pelvis as US scans of the abdomen, pelvis, and thyroid gland. In total, \textit{RadImageNet} constitutes a collection of 1.35 million radiologic images, offering a diverse and extensive resource for training one robust medical image generation diffusion model.

\begin{figure*}
\centering
\includegraphics[width=1.\textwidth]{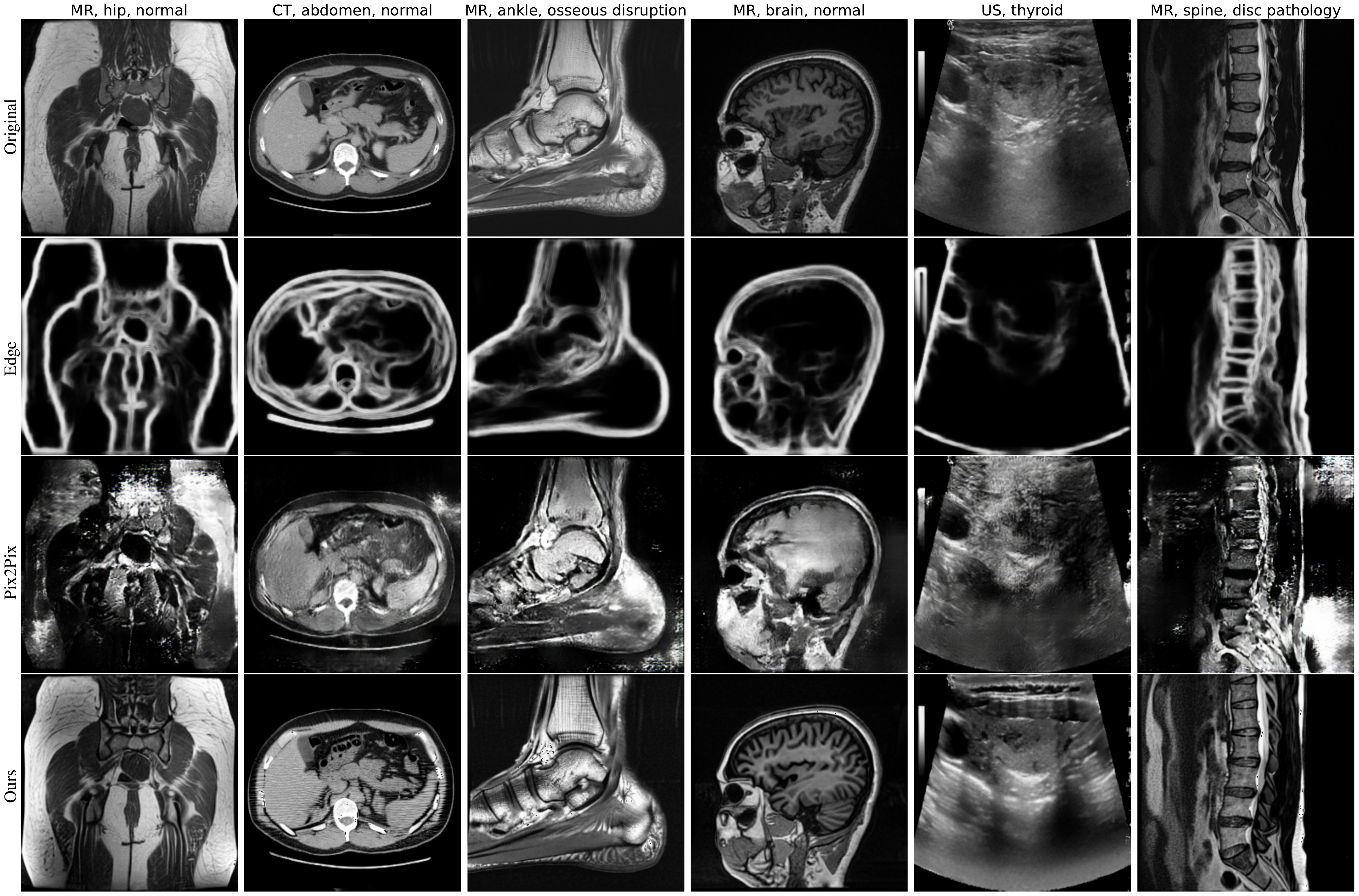}
\caption{\textcolor{black}{In the provided illustrations, the initial row presents example original images sourced from \textit{RadImageNet}. Middle row indicate edge maps of the original images to be used in diffusion process to enhance the visual quality. The last two rows includes sample images generated by the Pix2Pix GAN model and fine-tuned \textit{ControlNet} across diverse modalities. Both the original and synthesized images maintain congruent anatomical structures, albeit there may be disparities in intensity.} }
\label{fig:radimagenet}
\end{figure*}

In generating free training prompts using \textit{RadImageNet}, we employed a triplet: a combination of \textit{data modality}, \textit{organ name}, and \textit{category name}. For instance, a single CT scan image of the abdomen with arterial pathology would utilize prompts in the form of ``CT, Abdomen, Arterial Pathology." Similarly, a single MRI image of the hip with Chondral pathology would use the prompt of ``MR, Hip, Chondral pathology". This methodology ensures a systematic and consistent generation of free prompts across various medical imaging contexts while maintaining accurate medical terminology guiding precise medical structure generation. \textcolor{black}{As discussed earlier, our approach utilizes both text and edge conditioning inputs. To generate the edge conditioning input from \textit{RadImageNet}, we employ the Holistically-Nested Edge Detection (HED) algorithm proposed by~\cite{xie2015hed}. HED is a sophisticated deep learning-based edge detection technique that efficiently and accurately identifies image boundaries. By leveraging the extracted hidden features, we can generate precise edges not only for the skull but also for soft-tissue organs like the liver, spleen, or pancreas.} HED learns rich hierarchical representations under the guidance of deep supervision, rendering it a highly accurate tool for directing the generation of realistic and meaningful medical images.

\textbf{Training details:} We then adopt the stable diffusion model implementation with pre-training checkpoints in large-scale text-to-image datasets \cite{rombach2022stablediffusion,zhang2023conditiondiff}. With text and edge conditioning inputs, the diffusion model is trained on \textit{RadImageNet} using the AdamW optimizer. The training process is conducted utilizing 6 NVIDIA RTX A6000 GPUs, each equipped with 48GB memory and a batch size of 384 in total (48 for each under the DDP setting). The entire training procedure takes approximately seven days to complete, with a learning rate set at $10^{-5}$. We release all checkpoints and provide sample results from our proposed model in Figure~\ref{fig:radimagenet}, showing that real and generated images share similar anatomical information regardless of potential intensity differences.

\subsection{Finetuning on Downstream Task}
\textcolor{black}{While training a relatively large model offers advantages, it may not fully capture the intricacies of the data distribution specific to each medical application.} Particularly, the pre-trained model might lack the capability to generate specific segmentation targets, which is desired in the segmentation task. To address this limitation and increase the model's adaptability to diverse medical tasks, we fine-tune the pre-trained large-scale text-to-image medical generation stable diffusion model. \textcolor{black}{The fine-tuning process enables the model to learn task-specific features and variations relevant to the segmentation task at hand. This tailored approach leads to the generation of more pertinent synthetic samples for data augmentation, consequently enhancing the overall performance.} This process not only enhances the applicability of the model across a wide range of medical tasks but also ensures a higher degree of consistency and accuracy in the generated samples. As a result, the fine-tuned diffusion model is better equipped to contribute to more reliable generated augmentation data and improved performance and generalization in various medical image analysis tasks.

\textbf{Training details:} During the fine-tuning stages, we (continue to) incorporate both text and edge conditioning inputs to ensure accurate medical image generation. In contrast to the pre-training on \textit{RadImageNet}, we specifically incorporate the edge derived from the segmentation mask into the generation condition here. This approach is taken to ensure the accurate representation of anatomical structures for segmentation targets. Leveraging the robust initial starting point obtained from the pre-trained model, we utilize a single NVIDIA RTX A6000 GPU for each subtask training, maintaining a batch size of 48 while employing the AdamW optimizer. The learning rate is set at $10^{-6}$, and the fine-tuning process is performed over 100 epochs on the training set. To prevent potential data leakage, we exclusively use image-text pairs from the training set for fine-tuning the generation diffusion model. By adopting this cautious approach, we minimize the risk of inadvertently incorporating information from validation or test sets, thereby ensuring a more reliable evaluation of the model's performance and generalization capabilities.
\begin{figure*}[h]
\centering
\includegraphics[width=1\textwidth]{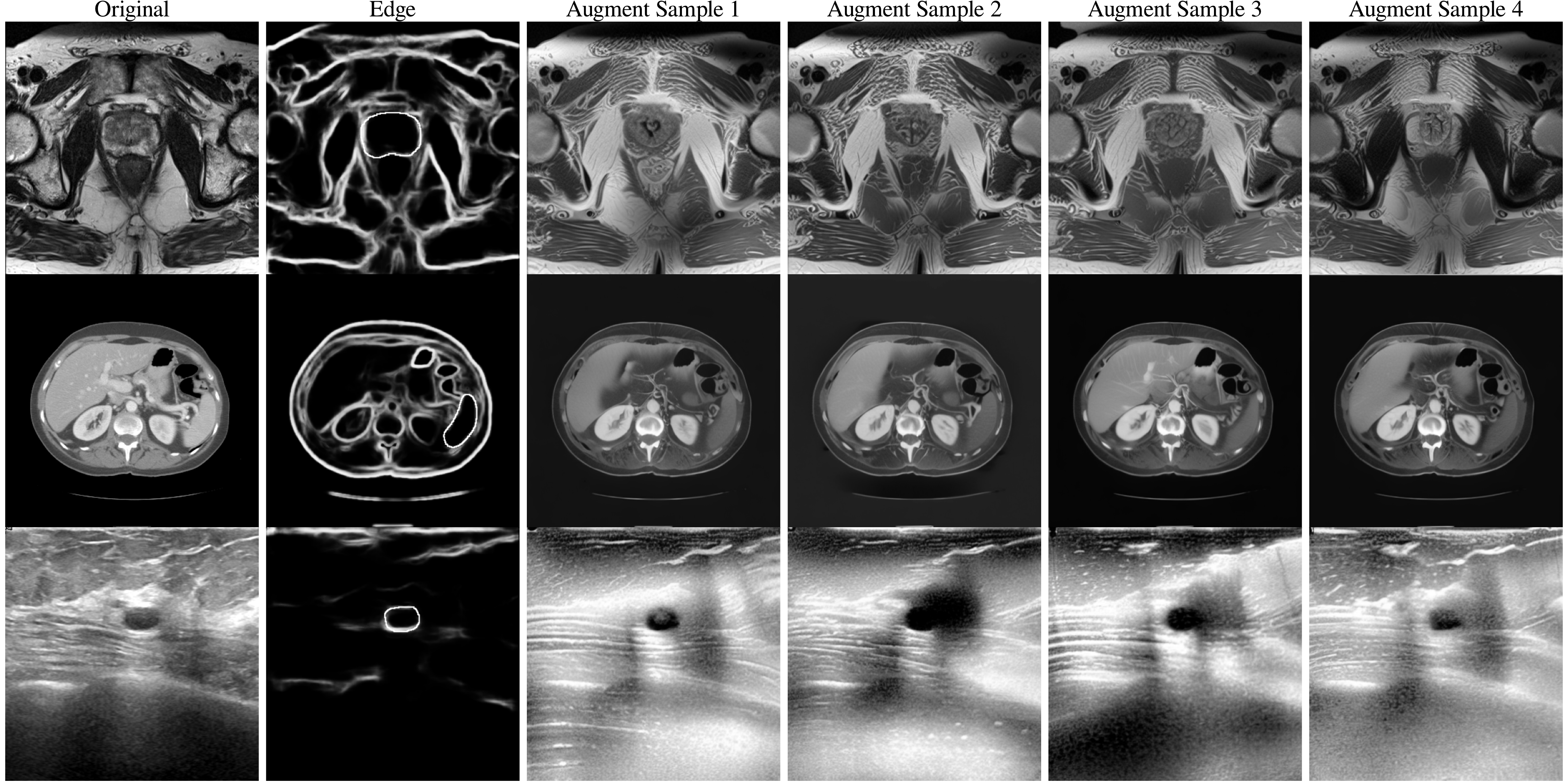}
\caption{Example augmented samples are illustrated in the last four columns while the first and second columns show original MRI, CT, and US images and their edge maps, respectively. Augmented samples show notable variances in intensity distribution (diversity) while they retain the structural integrity. }
\label{fig:generate_samples}
\end{figure*}

\subsection{Training of Downstream Task - Segmentation}
In this study, we primarily concentrated on segmentation tasks as a downstream application for assessing the efficacy of our data augmentation approach. The previous fine-tuned diffusion models generate new synthetic samples for data augmentation by adding augmentation text $c_{aug}$, like ``enhanced contrast", and ``high resolution," with the original text $c_t$ as conditioning input. By introducing these augmentation texts, we can generate more diverse data for promoting better generalization and performance. At the same time, these samples are created with text and edge information guidance to ensure that they represent the target distribution and exhibit meaningful attributes relevant to the downstream task. The generated synthetic samples were combined with the original training data during downstream task (segmentation) training. This integration allows the model to learn from both real and augmented samples, effectively increasing the diversity and volume of the training data. The choice of network architecture is not confined to the specific training procedure presented in this study, as our method is compatible with various segmentation loss designs.

\begin{algorithm}
  \caption{Training procedure for segmentation}
  \small
  \begin{algorithmic}[1]
    \vspace{.04in}
    \Statex Given a trained diffusion model $\mathbf{D}$, and $(\textbf{x}_0, c_{t}, c_{e}, \textbf{y})$ represents the image, text, edge, and label pairs in the training set
    \Statex Training target: any arbitrary segmentation network $\mathcal{C}$ with segmentation loss function $\textit{\textbf{l}}$ with loss balance hyper-parameter $\alpha$
    \State Augment $\textbf{x}_0$ by $n$ times $\epsilon\sim \mathcal{N}(0; 1)$ 
    \Statex \qquad $\textbf{x}_i = \mathbf{D}(\epsilon_i, c_{t} + c_{aug_i}, c_{e})$, \quad $i \sim 1, \dotsc, n$
    \For{$t=1, \dotsc, epochs$}
      \State Random choose $i \sim 1, \dotsc, n$ 
      \State $ \textit{\textbf{m}} = $ generate-random-patch$(\alpha, patch\ size)$
      \State $ \textit{\textbf{loss}} = \textit{\textbf{l}}(\mathcal{C}( \textbf{m} \cdot \textbf{x}_0 + (1-\textbf{m}) \cdot \textbf{x}_i), \textbf{y})$
      \State $\textit{\textbf{optimizer}}.zero\_grad()$
      \State $\textit{\textbf{loss}}.backward()$
      \State $\textit{\textbf{optimizer}}.step()$
    \EndFor
    \State \textbf{return} $\mathcal{C}$
    \vspace{.04in}
  \end{algorithmic}
  \label{alg:train_method}
\end{algorithm}

\begin{figure*}[h]
\centering
\includegraphics[width=1\textwidth]{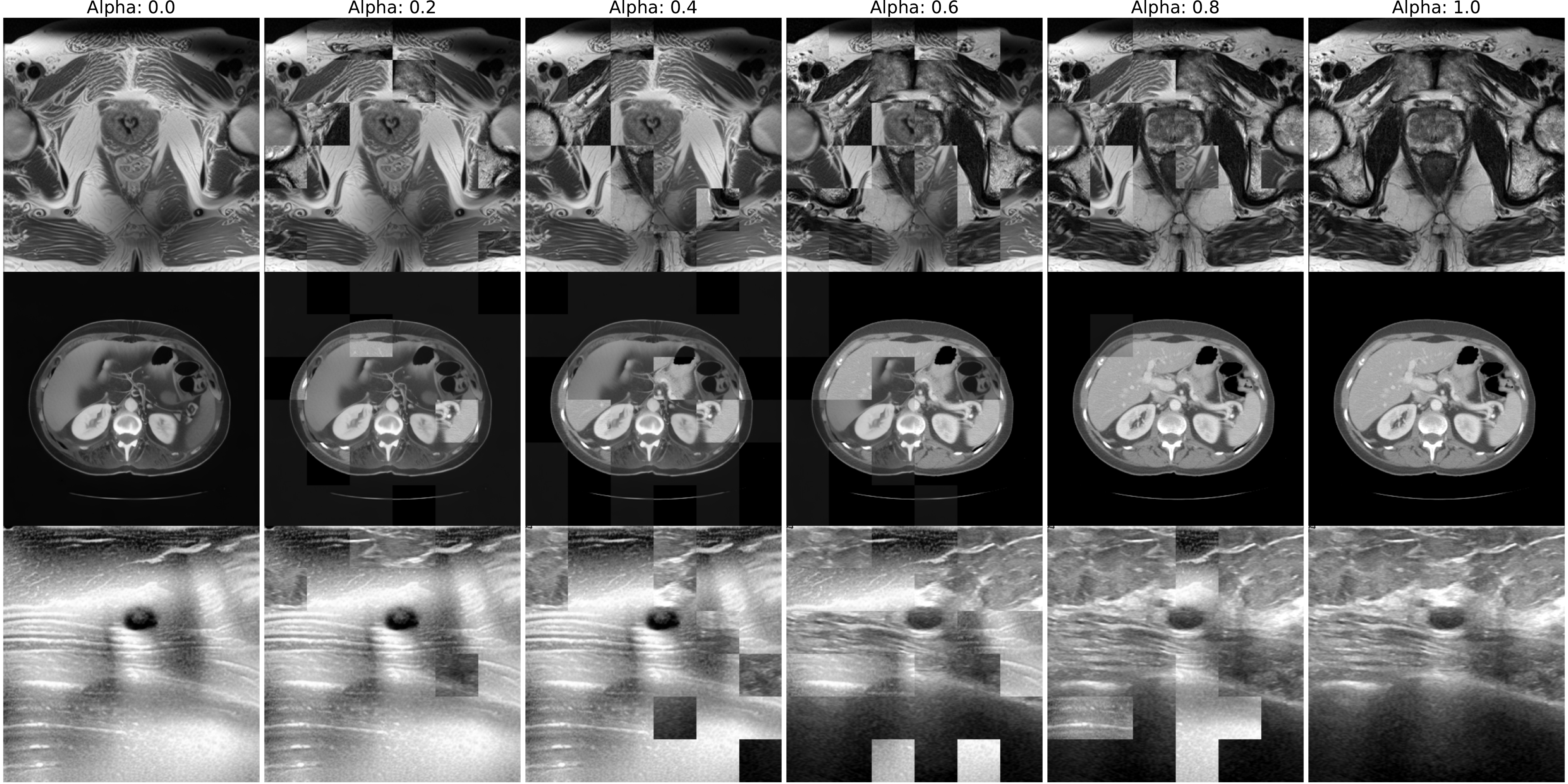}
\caption{\textcolor{black}{The hyper-parameter $\alpha$ determines the combination balance between original and augmented samples at the patch level.}}
\label{fig:visual_alpha}
\end{figure*}

\textcolor{black}{During the model training, we integrated both real and synthetic samples by employing a randomly generated patch mask for each image, determined by a hyper-parameter $\alpha$ and a pre-specified resolution. For instance, given a dimension of 384$\times$384, as is customary in our studies, and a patch size of 64, we would derive a random matrix with a dimension of 384/64$\times$384/64 from a uniform distribution ranging between 0 and 1. Every patch with values lower than a specific threshold $\alpha$ is converted to 1, while all others become 0. This simplified matrix is then resized back to its original dimensions using a nearest-neighbor interpolation technique. Intuitively, when $\alpha$ is large, more patches will be set as 1, and the combined (mixed) image will be more similar to the original samples, and when $\alpha$ is small, more patches will be set as 0 and the combined (mixed) image will be more similar to the generated samples. This procedure enables the fusion of real and synthetic images at the patch level. Thus, the hyperparameter $\alpha$ governs the mixing ratio, with higher values favoring the inclusion of real image patches and lower values incorporating more synthetic patches into the final representation. Figure~\ref{fig:visual_alpha} illustrates the influence of $\alpha$ including the extreme cases when $\alpha=0$ (meaning that patches are from generated samples only) nand $\alpha=1$ (meaning that patches are from original images only). Such strategy ensures that the model is not predisposed to genuine or artificial data, fostering improved generalization and resultant performance. We employed an ablation study to systematically investigate the impact of different hyperparameters  and image resolution choices on model performance(s). For a comprehensive understanding of the training algorithm itself, please refer to Algorithm~\ref{alg:train_method}.
}

\section{Results}
\subsection{{Diffusion Model for \textit{RadImageNet}}}

\textcolor{black}{The pre-trained diffusion model demonstrated a promising performance on the \textit{RadImageNet}  for data generation. We measured the quality of the generated synthetic images using the Mean Absolute Error (MAE), Mean Squared Error (MSE), Root Mean Square Error (RMSE), and MultiScale Structural Similarity Metrics (MS-SSIM). Prevalent metrics used in natural image generation, such as Frechet Inception Distance (FID) and Inception Score (IS), are underpinned by a feature extractor pre-trained on the ImageNet dataset. The inherent characteristics and complexities of medical images, which are fundamentally different from natural images, render these metrics less effective in accurately assessing the performance of our approach. }

\textcolor{black}{We compared our diffusion-based generation methods with the  widely adopted conditional GAN: Pix2Pix image translation~\cite{isola2017pix2pix}. Pix2Pix takes the same edge information as guidance and reconstructs the original input image. The diffusion model achieved an MAE score of 0.087, an MSE of 0.023, an RMSE of 0.144, an SSIM of 0.636, and an MS-SSIM of 0.636 while the Pix2Pix GAN method achieved an MAE score of 0.147, an MSE of 0.0538, an RMSE of 0.2219, an SSIM of 0.4583, and an MS-SSIM of 0.5325, as shown in Table~\ref{tab:radimagenet}, indicating a high fidelity and diversity of the generated medical images.} \textcolor{black}{These results suggest that the diffusion model is effective in generating realistic and meaningful medical images, which in turn can enhance the performance of downstream medical image analysis tasks.} \textcolor{black}{Figure}~\ref{fig:generate_samples} \textcolor{black}{indicates several sampling results when combined with text-guidance.}

\begin{table*}[h]
\caption{\textcolor{black}{Performance on the \textit{RadImageNet} dataset generation. Data range is between 0-1 for computation.}}
\centering
\begin{tabular}{c|c|c|c|c|c}
\hline
\hline
 & MAE $\downarrow$ & MSE $\downarrow$ & RMSE $\downarrow$ & SSIM $\uparrow$ & MS-SSIM $\uparrow$ \\
\hline
Pix2Pix~\cite{isola2017pix2pix} & 0.1469 ± 0.0556 & 0.0538 ± 0.0366 & 0.2219 ± 0.0676 & 0.4583 ± 0.1105 &0.5325 ± 0.1557\\
\hline
Ours & 0.0873 ± 0.0363 & 0.0229 ± 0.0163 & 0.1441 ± 0.0463  & 0.6356 ± 0.1252 & 0.6666 ± 0.1491 
\\
\hline
\hline
\end{tabular}
\label{tab:radimagenet}
\end{table*}


\textcolor{black}{It should be noted that our focus in this study is on generating controlled, clinically relevant variations with text prompts.} \textcolor{black}{In other words, our work aims to strike a balance between the two (segmentation and data diversity), not maximizing one over another. One may conclude that Pix2Pix may generate more diverse images regarding MS-SSIM. However, it should be noted that Pix2Pix does not take text prompts as input and can hardly be used to generate more diverse data with semantic diversity for enhancing segmentation performance. The MS-SSIM metric primarily measures pixel-level similarity, which can be high even when images exhibit meaningful semantic variations. Our model’s higher MS-SSIM scores indicate better structural consistency with the ground truth, not necessarily less meaningful diversity.}

\subsection{{Performance on Downstream Tasks}}

\begin{table*}[!h]
\caption{Comparing with other data augmentation methods on Ultrasound, CT, and MRI modalities across various organs, including breast cancer, spleen, and prostate segmentation datasets. We can observe superior performance of the \textcolor{black}{DiffBoost} technique over other data augmentation methods.}
\centering
\resizebox{0.8\textwidth}{!}{%
\begin{tabular}{c|l l l | l l}
\hline
\hline
Task & \multicolumn{4}{c}{Task 1: Prostate MRI Segmentation (Sample Size: 32 )} \\
\hline
Method & Dice $\uparrow$ & Precision $\uparrow$ & Recall $\uparrow$ & HD95 (mm) $\downarrow$ & ASSD (mm) $\downarrow$ \\
\hline
Baseline             &78.46 ± 10.36 &      77.72 ± 11.89 &   81.35 ± 13.11 &                   11.93 ± 8.04 &                              3.12 ± 1.32 \\
RandomContrast       &81.23 ± 8.79 &      82.13 ± 10.25 &    82.10 ± 12.07 &                   10.41 ± 6.31 &                               2.59 ± 1.00\\
RandomGamma          &79.30 ± 9.05 &      79.37 ± 11.15 &   82.17 ± 11.98 &                   11.65 ± 7.07 &                              2.84 ± 1.44 \\
RandomBrightness     &79.18 ± 8.07 &      80.97 ± 10.02 &   79.81 ± 12.54 &                    11.10 ± 9.07 &                              2.91 ± 1.18 \\
RandomNoise          &80.68 ± 7.29 &      80.34 ± 10.38 &   82.55 ± 10.33 &                   12.04 ± 9.43 &                              2.68 ± 1.06 \\
RandomResolution     &78.63 ± 8.54 &      80.82 ± 10.85 &   78.79 ± 12.56 &                  12.22 ± 11.25 &                              2.85 ± 1.33 \\
RandomMirror         &79.23 ± 11.32 &       85.08 ± 9.93 &   76.19 ± 15.94 &                    9.31 ± 5.46 &                              2.55 ± 1.18 \\
RandomRotate         &82.12 ± 9.03 &       85.62 ± 8.33 &   80.53 ± 12.67 &                    8.47 ± 7.47 &                               2.30 ± 1.39 \\
RandomScale          &83.11 ± 7.29 &       84.59 ± 9.26 &   83.18 ± 10.53 &                    9.64 ± 7.84 &                              2.37 ± 1.06 \\
DeepStack            &78.97 ± 9.40 &      81.86 ± 10.21 &   78.99 ± 14.33 &                    9.32 ± 4.61 &                               2.7 ± 1.17 \\ \hline
\textcolor{black}{DiffBoost} (ours) &\textbf{84.56 ± 6.69} &         \textbf{86.70 ± 6.50} &  \textbf{ 83.98 ± 10.53} &     \textbf{ 7.75 ± 8.88} &                   \textbf{2.06 ± 1.40} \\ \hline
\hline
Task & \multicolumn{4}{c}{Task 2: Spleen CT Segmentation  (Sample Size: 41) } \\
\hline
Method & Dice $\uparrow$ & Precision $\uparrow$ & Recall $\uparrow$ & HD95 (mm) $\downarrow$ & ASSD (mm) $\downarrow$ \\
\hline
Baseline             &  94.42 ± 2.76 &     95.12 ± 2.84 &  93.92 ± 4.67 &                  5.18 ± 4.43 &                            0.94 ± 0.68 \\
RandomContrast       &  94.29 ± 2.55 &     95.15 ± 3.43 &  93.66 ± 4.29 &                 8.18 ± 12.58 &                            1.35 ± 1.38 \\
RandomGamma          &  94.69 ± 1.98 &      95.37 ± 2.80 &  94.15 ± 3.31 &                  5.59 ± 5.42 &                            1.14 ± 1.43 \\
RandomBrightness     &  93.84 ± 2.82 &     94.17 ± 3.83 &  93.77 ± 4.68 &                  6.57 ± 6.04 &                            1.23 ± 1.04 \\
RandomNoise          &  93.84 ± 3.28 &     94.59 ± 3.71 &  93.44 ± 5.53 &                  6.32 ± 5.15 &                            1.29 ± 0.94 \\
RandomResolution     &  93.99 ± 3.01 &     94.66 ± 4.22 &  93.68 ± 5.04 &                  7.84 ± 7.74 &                            1.29 ± 1.12 \\
RandomMirror         &  91.95 ± 4.89 &     92.92 ± 5.05 &  91.52 ± 7.46 &                22.89 ± 38.69 &                            3.23 ± 3.30 \\
RandomRotate         &  93.76 ± 4.46 &    \textbf{95.49 ± 3.17} &   92.6 ± 7.65 &         6.15 ± 6.27 &                              1.17 ± 1.01 \\
RandomScale          &  93.98 ± 2.65 &     94.82 ± 4.11 &  93.44 ± 4.21 &                  5.96 ± 5.39 &                            1.09 ± 0.81 \\
DeepStack            &  93.66 ± 3.36 &      93.73 ± 4.60 &  93.99 ± 5.55 &                   8.3 ± 8.28 &                           1.39 ± 1.15 \\ \hline
\textcolor{black}{DiffBoost} (ours) &  \textbf{94.78 ± 1.95} &94.14 ± 3.13 &\textbf{95.55 ± 2.68} &\textbf{4.88 ± 3.83} &\textbf{0.9 ± 0.55} \\
\hline
\hline
Task & \multicolumn{4}{c}{Task 3: Breast Cancer Ultrasound Segmentation  (Sample Size: 147)} \\
\hline
Method & Dice $\uparrow$ & Precision $\uparrow$ & Recall $\uparrow$ & HD95 (pixel) $\downarrow$ & ASSD (pixel) $\downarrow$ \\
\hline
Baseline             &      62.92 ± 25.79 &           63.34 ± 31.15 &         76.59 ± 24.50 &                     138.19 ± 118.30 &                                 38.20 ± 38.20 \\
RandomContrast       &      64.81 ± 27.36 &           65.83 ± 30.89 &        73.93 ± 27.96 &                     123.24 ± 105.32 &                                 34.96 ± 33.83 \\
RandomGamma          &      66.43 ± 25.11 &           67.66 ± 29.26 &         75.40 ± 25.51 &                     117.96 ± 113.57 &                                 33.63 ± 34.13 \\
RandomBrightness     &      63.65 ± 27.58 &           65.56 ± 31.79 &        73.42 ± 26.95 &                     125.58 ± 111.85 &                                 37.63 ± 39.78 \\
RandomNoise          &      66.24 ± 25.52 &           67.29 ± 30.73 &        75.79 ± 22.92 &                     132.05 ± 120.47 &                                 35.65 ± 34.14 \\
RandomResolution     &      67.89 ± 25.64 &           68.15 ± 29.01 &        77.03 ± 23.88 &                     127.33 ± 124.93 &                                 34.42 ± 37.11 \\
RandomMirror         &      70.15 ± 25.55 &           \textbf{70.85 ± 28.77} &78.45 ± 25.14 &                     102.08 ± 105.71 &                                 29.83 ± 36.53 \\
RandomRotate         &      69.15 ± 27.57 &           69.99 ± 30.95 &         77.87 ± 25.20 &                     118.18 ± 129.85 &                                 34.33 ± 44.07 \\
RandomScale          &      69.15 ± 26.05 &           68.96 ± 29.15 &        79.24 ± 25.84 &                      113.74 ± 117.50 &                                 32.01 ± 36.97 \\
DeepStack            &      63.85 ± 25.63 &           63.64 ± 30.05 &        76.33 ± 23.94 &                     144.79 ± 120.06 &                                 37.94 ± 38.56 \\ \hline
\textcolor{black}{DiffBoost} (ours)    &\textbf{71.65 ± 24.52} &    70.22 ± 27.89 &\textbf{81.91 ± 22.39} &\textbf{95.18 ± 103.41} &\textbf{26.98 ± 32.55} \\
\hline
\hline
\end{tabular}}
\label{tab:main_result}
\end{table*}

\textcolor{black}{We validate our model's performance on multiple datasets with limited sample sizes, including ultrasound, CT, and MRI modalities across various organs like breast ~\cite{al2020breast}, spleen~\cite{antonelli2022msd}, and prostate~\cite{antonelli2022msd}. For 3D datasets like CT spleen or MRI prostate, we split them into 2D with a slice depth of one. We use the standard AttentionUNet~\cite{oktay2018attentionunet} as the segmentation backbone with MONAI implementation~\cite{cardoso2022monai}, and in the ablation studies, we investigate the combination effects between} \textcolor{black}{DiffBoost} \textcolor{black}{and various segmentation backbones. For comparison, we investigated different traditional medical augmentation methods, including spatial transforms like Random Rotate, Random Scale, and Random Mirror, Random Resolution, and intensity transform like Random Contrast, Random Gamma, Random Brightness, Random Noise, and the Deep Stack transform,} \textcolor{black}{which indicates the combination of all previous transforms implemented in nnUNet~\cite{isensee2021nnunet}}. \textcolor{black}{We have chosen AdamW as optimizer, and all experiments were conducted under 3-fold cross-validation. Model performance was comprehensively assessed using two categories of metrics: region-level metrics such as Dice coefficient (Dice), Precision, and Recall, and shape-centric metrics like the 95\% Hausdorff Distance (HD95) and Average Symmetric Surface Distance (ASSD). This dual-metric approach facilitates a detailed and rigorous evaluation of our model's capabilities.}

\textcolor{black}{Table~\ref{tab:main_result} demonstrates the superiority of DiffBoost compared to other data augmentation techniques. Notably, DiffBoost achieves significant improvements in Dice coefficient, suggesting that DiffBoost effectively guides the model towards learning more robust and intensity-independent features, particularly the morphology (shape and structure) of the target organs.} \textcolor{black}{Beyond the average performance boost, DiffBoost also leads to a reduction in standard deviation across datasets. This decreased variability implies that the model trained with DiffBoost captures features more consistently. This robustness is further supported by the observed reduction in HD95 distance across datasets. This metric measures the average distance between the predicted segmentation boundary and the ground truth, indicating that DiffBoost generates segmentation masks with more precise and consistent shapes for anatomical structures. These findings highlight DiffBoost's potential as a powerful tool for achieving robust medical image segmentation, particularly in tasks where consistent and accurate identification of anatomical structures is critical.}


\textcolor{black}{DiffBoost demonstrates significant improvements in segmentation performance, particularly for challenging tasks like prostate MRI and breast ultrasound segmentation. The Dice coefficient increases from 78.46\% to 84.56\% (7.8\% improvement) for prostate MRI and from 62.92\% to 71.65\% (13.87\% improvement) for breast cancer segmentation. These results highlight DiffBoost's ability to enhance model performance in complex segmentation scenarios where structural information is crucial. For less complex tasks like spleen CT segmentation, where the baseline approach already achieves a high accuracy (Dice coefficient of 94.42\%), the improvement from DiffBoost was marginal (94.78\%). This suggests that for tasks with well-defined features and high baseline performance, data augmentation methods may have a less pronounced effect. It's important to note that combining multiple data augmentation techniques (like \textit{DeepStack}) doesn't always guarantee better performance compared to a single, well-designed approach (as shown in previous works~\cite{cubuk2019autoaugment,cubuk2020randaugment,muller2021trivialaugment}). DiffBoost stands out by consistently delivering noteworthy advancements even for tasks with already strong baseline performance.}


\begin{figure*}
\centering
\includegraphics[width=1.\textwidth]{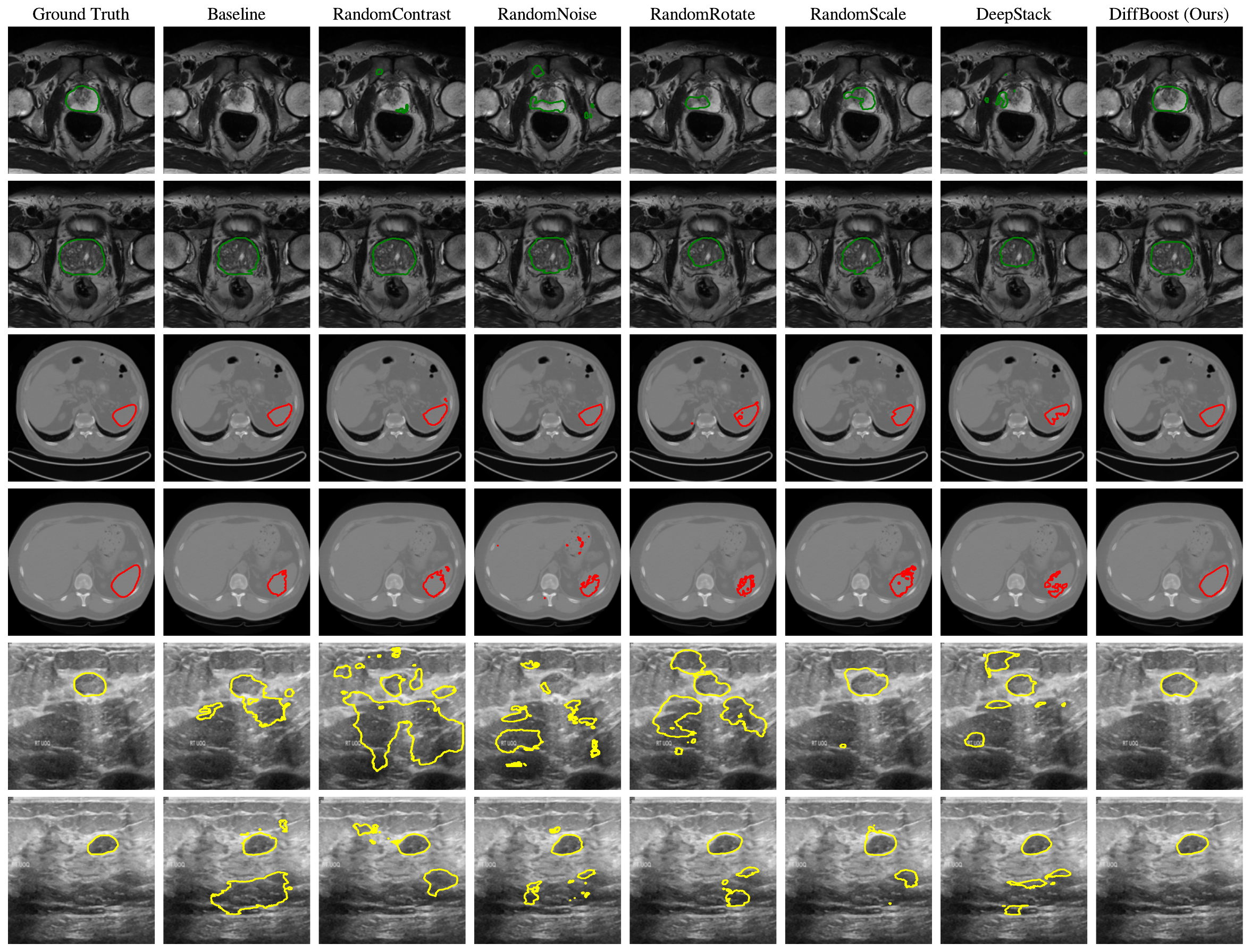}
\caption{ \textcolor{black}{Visual comparison of segmentation performance over some other augmentation methods. DiffBoost enables the model to generate a more consistent shape of the anatomical structure and outperforms other data augmentation methods.}}
\label{fig:segvisual}
\end{figure*}

\textcolor{black}{Beyond the quantitative comparisons, we conducted a qualitative analysis (Figure~\ref{fig:segvisual}) to visually assess the performance of our method. Interestingly, DiffBoost leads to a noticeable improvement in segmentation reliability, especially for challenging tasks like ultrasound breast cancer segmentation. This can be attributed to the model's focus on structural information beyond just intensity distribution. As a result, the model excels at delineating suspicious regions with greater accuracy. These qualitative findings strongly support the effectiveness of DiffBoost as a data augmentation approach. By generating meaningful and realistic synthetic medical images, DiffBoost can significantly improve the performance of downstream medical image segmentation tasks.}

\section{Ablation Experiments}
\textcolor{black}{We conducted extensive ablation experiments to investigate the influence of different data augmentation ratios, hyper-parameter settings, patch size settings, and conjunction effects with various network architectures on the prostate MRI segmentation dataset.}

\begin{figure*}[!t]
\centering
\includegraphics[width=0.99\textwidth]{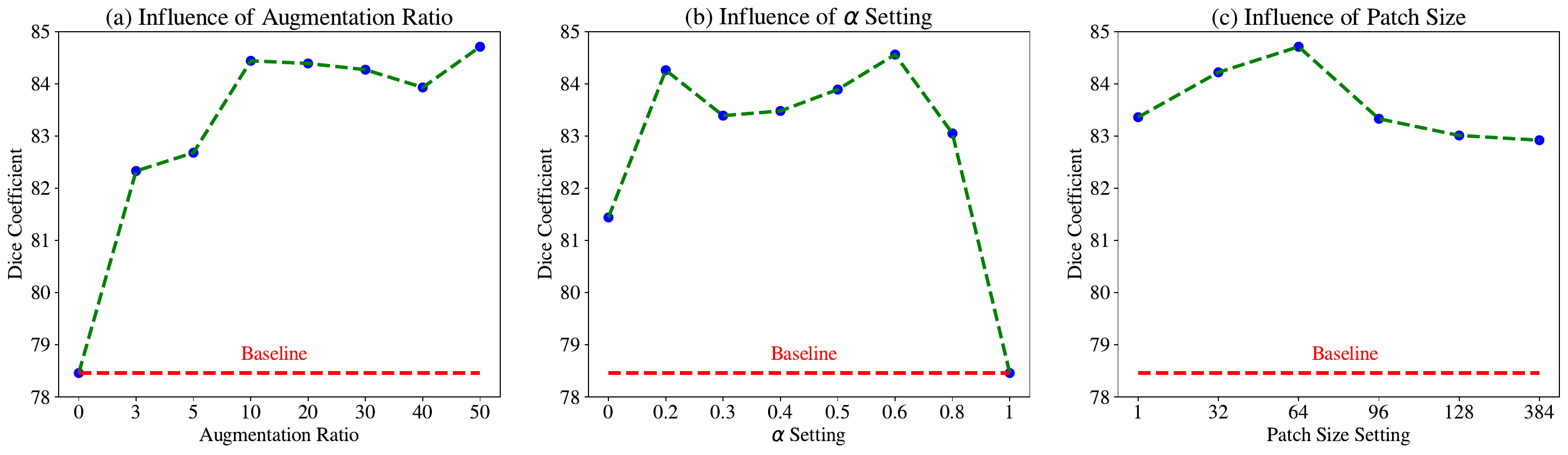}
\caption{(a) Influence of data augmentation ratios on segmentation performance. The segmentation performance increases significantly as the data augmentation ratio increases. The marginal performance gain is limited once the ratio is larger than 10. (b) Influence of hyper-parameter $\alpha$ setting on segmentation performance. The optimal performance of the model can be adversely affected by the settings for $\alpha$ if it is either excessively high or low. A high-performance plateau exists within the wide middle range. (c) Influence of patch size on segmentation performance. Regardless of patch size setting, \textcolor{black}{DiffBoost} can achieve significant advancement over the baseline method. }
\label{fig:ablation_study}
\end{figure*}

\subsection{Influence of Augmentation Ratio}

\textcolor{black}{The data augmentation ratio, defined as $n$ in Algorithm ~\ref{alg:train_method} for how many samples we generated from the original images, can have a notable impact on the final performance of the model. As one might expect, a higher data augmentation ratio can lead to more robust performance, although the marginal gains may diminish as the ratio increases. A higher augmentation ratio can also increase computational demands, which is undesirable for diffusion models due to their already resource-intensive nature. Therefore, it is crucial to strike an appropriate balance between the data augmentation ratio and computational efficiency when employing diffusion models for medical imaging applications. As shown in  Figure~\ref{fig:ablation_study}(a), observations gleaned from our analysis indicate a notable performance increase when the data augmentation ratio is relatively small. However, when this ratio exceeds a factor of 10, the marginal performance enhancements begin to plateau, exhibiting a limited increase and some degree of fluctuation. This suggests that a higher data augmentation ratio does not necessarily translate to proportionally larger performance improvements, and the marginal performance gain might be limited compared with a large computation burden.}

\subsection{Influence of Hyper-parameter $\alpha$}
\textcolor{black}{The hyper-parameter $\alpha$ plays a significant role in our framework, enabling us to control the balance between the contributions of real training samples and those synthetically generated by our diffusion model. Specifically, $\alpha$ acts as a weighting factor in our pre-designed combination loss function, where a larger $\alpha$ assigns more importance to the real samples, and a smaller $\alpha$ increases the influence of the generated samples. In this ablation study, we explore the impact of various hyper-parameters on the final performance metrics. As expected, when $\alpha$ is set exceedingly high, the model leans towards the original samples, resulting in a marginal increase in performance. Conversely, an excessively low $\alpha$ would cause the model to overly concentrate on synthetic samples, adversely affecting performance due to the lack of guidance from real samples. This anticipated trend is explicitly corroborated by Figure~\ref{fig:ablation_study} (b), which presents a high-performance plateau when the data augmentation ratio resides within the wide range of 0.2-0.8.}

\subsection{Influence of Patch Size}
\textcolor{black}{The choice of patch size is critical in our experiments as it dictates the spatial granularity at which real and synthetic samples are combined. In cases where the patch size matches the image size, the selection between real and synthetic samples occurs on a case-by-case basis. Conversely, when the patch size is set to 1, selection occurs at the pixel level. Empirical findings in Figure~\ref{fig:ablation_study} (c) suggest that the} \textcolor{black}{DiffBoost} \textcolor{black}{significantly improves over the baselines, irrespective of the patch size setting. Moreover, optimal performance was noted when the patch size was set to a median spatial level, illustrating the importance of carefully balancing fine-grained and broad spatial representations.}

\subsection{Influence of Network Architecture}
\begin{figure}
\centering
\includegraphics[width=0.49\textwidth]{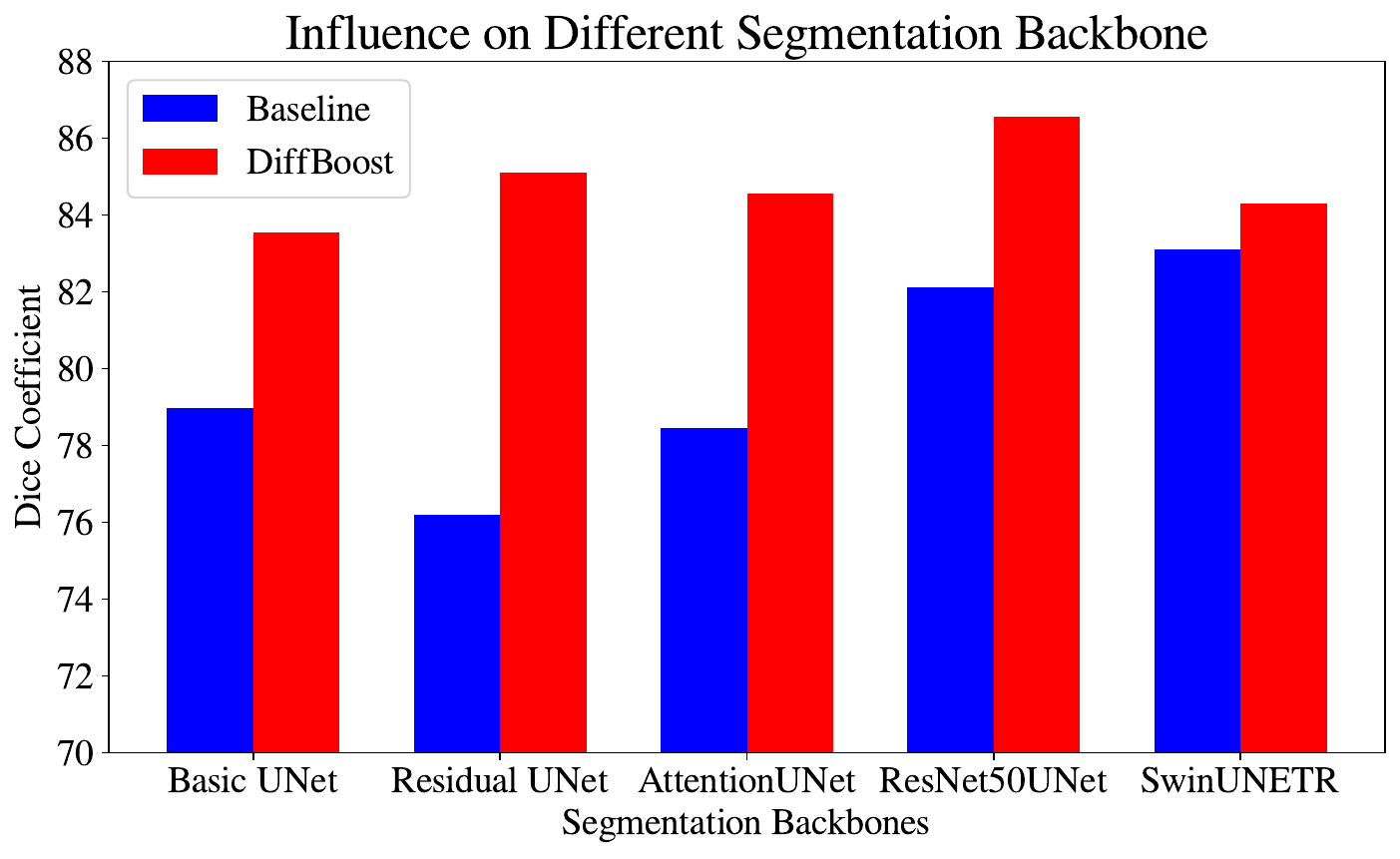}
\caption{Influence of backbone architectures on segmentation results. \textcolor{black}{DiffBoost} achieves a promising and consistent improvement regardless of the choice of different backbones.}
\label{fig:segmentation_backbone}
\end{figure}

\textcolor{black}{To ensure the general applicability of proposed} \textcolor{black}{DiffBoost,} \textcolor{black}{it is important to investigate its performance in conjunction with various network architectures, as structural differences may impact the final performance. To this end, we conducted an ablation study to assess the combined influence of different network architectures and the} \textcolor{black}{DiffBoost} \textcolor{black}{method on the prostate MRI segmentation tasks. Besides the backbone AttentionUNet in the previous section, we include traditional CNN structures like basic UNet~\cite{falk2019unet}, Residual UNet~\cite{kerfoot2019resunet}, ResNet50 UNet~\cite{he2016dresnet50}, and the recent transformer structure like SwinUNETR~\cite{hatamizadeh2021swinunet}. This comprehensive analysis, as shown in Figure~\ref{fig:segmentation_backbone}, demonstrates the robustness and adaptability of} \textcolor{black}{DiffBoost} \textcolor{black}{across a diverse range of architectural designs, further supporting its utility in a wide array of medical imaging segmentation applications.} 

\section{Discussion and Conclusion}
\textcolor{black}{In this study, we proposed a diffusion model-based data augmentation approach, called \textcolor{black}{DiffBoost}, for enhancing the performance of medical image segmentation. The method demonstrated promising results in generating  anatomy-meaningful synthetic images conditioned on text and edge information.} 

\textcolor{black}{Our study has some limitations too. For instance, our current text input is a combination of several categorical labels. We have not tested how the model will perform under natural texts, like \textit{"Can you help to generate a CT image of a spleen."}, and how the performance will differ compared to categorical label based generation. We envision that it should be possible to generate similar results under natural text input too because the recent surge in techniques for aligning medical text data with large-scale models presents a unique opportunity allowing for the utilization of more natural language descriptions when generating medical images. In some cases, like in our study, simple categorical conditions or other input forms may already suffice to guide the segmentation process effectively. On the other hand, natural text input can be more useful when a segmentation task requires specific instructions and provides much more flexible input.}

\textcolor{black}{Another limitation is that employing the edge map (ControlNet) as a condition on image generation can limit the anatomical variation; hence, influencing the characteristics of the generated images. However, ControlNet can also be viewed as providing boundary information derived from the generated \textit{texture}. If ControlNet overly emphasizes or constrains certain features, there might occur a reduction in diversity in generated samples, affecting the segmentation results. This is not the case in our experiments, luckily, reflecting the fact that ControlNet is not excessively limiting the diversity, or anatomical variations  in the generated samples, thanks to parameter control in the edge locations in hierarchical ways. This potential limitation can be substantial if true edges are used instead of \textit{edgemaps}. It is because edgemaps are used at different sensitivity levels while it is hard to control the sensitivity of the true edges for different images, resolutions, and body regions.} 

\textcolor{black}{One may wonder if higher-fidelity synthetic images (than those already generated by our method) are truly necessary for successful segmentation. Based on our findings and limited studies in the literature, the answer is "no." High-fidelity synthetic scans are not always necessary for successful medical image segmentation. High accuracy can still be obtained with lower-fidelity synthetic images as long as latent space features are informative. However, we also acknowledge that higher-quality synthetic images can boost the segmentation results further. It should also be noted that the optimal approach for generating high-fidelity images will depend on various factors, including the complexity of the anatomy and the availability of large-scale real-world data. } 

\textcolor{black}{From the visual results, we observed mixed facts about the quality of boundary information in soft and bone tissue locations. For instance,  as shown in Figure~\ref{fig:radimagenet}, the skull in MRI can lead to clean edge guidance. Similarly, generated images from the text prompt \textit{“CT, abdomen, normal”} provide clear boundaries even from soft tissues (i.e., liver, pancreas, and spleen) due to the involvement of Holistically Nested edge detection method. We have observed the same behavior in prostate MRI too (as shown in Figure~\ref{fig:generate_samples}). However, there were cases where the soft tissue boundaries were not clear in the generated images. As a result, we did not have a clear consensus whether soft tissue generation is always inferior to other tissues or not.}

\textcolor{black}{Other than pix2pix, we did not benchmark other GAN-based methods like StyleGAN~\cite{karras2019stylegan} because our study primarily underscores the feasibility of diffusion model-based data augmentation for segmentation tasks; the complexity of tailoring these techniques to each medical dataset in our problem is computationally prohibitive and digressing from the main focus of this paper. Nevertheless, comprehensive comparison and analysis of generative AI algorithms from three broad classes  (GANs, VAEs, Diffusions) can be studied as a complementary to our current study. Yet, this does not undermine our core findings.}

\textcolor{black}{Our immediate future work will focus on (i) exploring techniques to accelerate the sampling process within the diffusion model for improved efficiency, (ii) investigating methods to incorporate text information more powerfully, allowing for even finer control over the generated images, and (iii) expanding image patterns beyond edge information to integrate broader contextual information from the text descriptions, potentially leading to richer and more nuanced synthetic images.}


 \bibliographystyle{IEEEtran}
\textcolor{black}{\bibliography{refs}}

\end{document}